\documentclass[11pt]{article}

\usepackage[final]{acl}

\usepackage{times}
\usepackage{latexsym}
\usepackage{xcolor}
\usepackage{balance}
\usepackage[T1]{fontenc}

\usepackage[utf8]{inputenc}

\usepackage{microtype}

\usepackage{inconsolata}

\usepackage{graphicx}

\usepackage{enumitem}
\usepackage{microtype}
\usepackage{graphicx}
\usepackage{subcaption}
\usepackage{booktabs} 

\usepackage{amsmath}

\usepackage{xcolor}
\usepackage{tcolorbox}
\tcbuselibrary{skins, breakable}
\newtcolorbox{reasonbox}[2]{
  enhanced,
  colback=#1!5,
  colframe=#1!60!black,
  boxrule=0.8pt,
  arc=2mm,
  left=6pt,
  right=6pt,
  top=6pt,
  bottom=6pt,
  title=\textbf{#2},
  fonttitle=\small\bfseries,
}

\newtcolorbox{problembox}{
  enhanced,
  colback=gray!5,
  colframe=gray!70!black,
  boxrule=0.8pt,
  arc=2mm,
  left=6pt,
  right=6pt,
  top=6pt,
  bottom=6pt,
}

\usepackage{tikz}
\usetikzlibrary{arrows.meta, positioning}

\usepackage{hyperref}
\usepackage{amssymb}

\usepackage{algorithm}
\usepackage{algpseudocode}
\usepackage{ulem}

\usepackage{amsmath}
\usepackage{amssymb}
\usepackage{mathtools}
\usepackage{amsthm}

\usepackage[capitalize,noabbrev]{cleveref}

\theoremstyle{plain}

\theoremstyle{definition}

\theoremstyle{remark}

\usepackage[textsize=tiny]{todonotes}

\title{Sampling via Decision-Flow: Training-Free Extraction of Improved Latent Reasoning Paths in Large Language Models}

\author{
 \textbf{Zhendong Mi},
 \textbf{Shaoyi Huang$^\dagger$}
\\
 \textsuperscript{}Stevens Institute of Technology
\\
   \texttt{\{zmi2, shuang59\}@stevens.edu}
}

\begin{document}
\maketitle
\begin{abstract}

A central question in LLM reasoning is whether reinforcement learning (RL) instills genuinely new capabilities or merely reshapes how existing knowledge is expressed during inference. Building on the distribution-sharpening hypothesis, which holds that RL reallocates probability mass toward high-reward trajectories already latent in base models, we ask: can we unlock those latent paths without costly RL fine-tuning? We present \textbf{Decision-Flow Sampling (DF-Sample)}, a training-free, data-free inference-time framework that constructs a hierarchical reasoning tree, scores terminal nodes for quality, and back-propagates utilities to inform each intermediate branching decision. Unlike conventional sampling strategies that make purely local step-wise choices, DF-Sample performs explicit global trajectory evaluation before committing to a path, recovering high-quality but low-probability reasoning chains that standard decoding overlooks. On GPQA, DF-Sample achieves 45.6\% accuracy, surpassing power sampling (38.9\%) and GRPO (39.9\%), showing that a training-free method can outperform a trained one. Across three models and four benchmarks, DF-Sample consistently outperforms baselines, indicating substantial latent reasoning potential in pretrained base models.

\end{abstract}

\section{Introduction}

Reinforcement learning has become the dominant post-training paradigm for improving LLM reasoning~\cite{guo2025deepseek,hu2025open}. 
By optimizing with outcome- or process-level rewards, RL-based post-training has pushed state-of-the-art performance on challenging benchmarks in mathematics, code, and science ~\cite{hendrycks2021measuring,li2022competition,rein2024gpqa}. 
Yet how RL improves reasoning remains actively debated
\cite{he2025rewarding,karan2025reasoning}. 

The \textit{Distribution Sharpening} hypothesis~\cite{shao2025spurious,zhang2025survey,yue2025does} provides one compelling answer: RL does not introduce genuinely new competencies, but rather reallocates probability mass toward high-reward trajectories that were already latent in the base model. While this effectively improves single-attempt metrics such as Pass@1, it simultaneously reduces output diversity by concentrating mass on high-confidence paths and suppressing alternative strategies.
This reduction in diversity comes at a cost. Several recent works have implicated it in a form of model collapse~\cite{he2025rewarding,hao2025rethinking,song2025outcome}, in which the model's reasoning coverage contracts even as peak performance improves \cite{liang2025beyond}. The model becomes more confident but less exploratory.

\begin{figure*}[!t]
\centering
\includegraphics[width=0.99\linewidth]{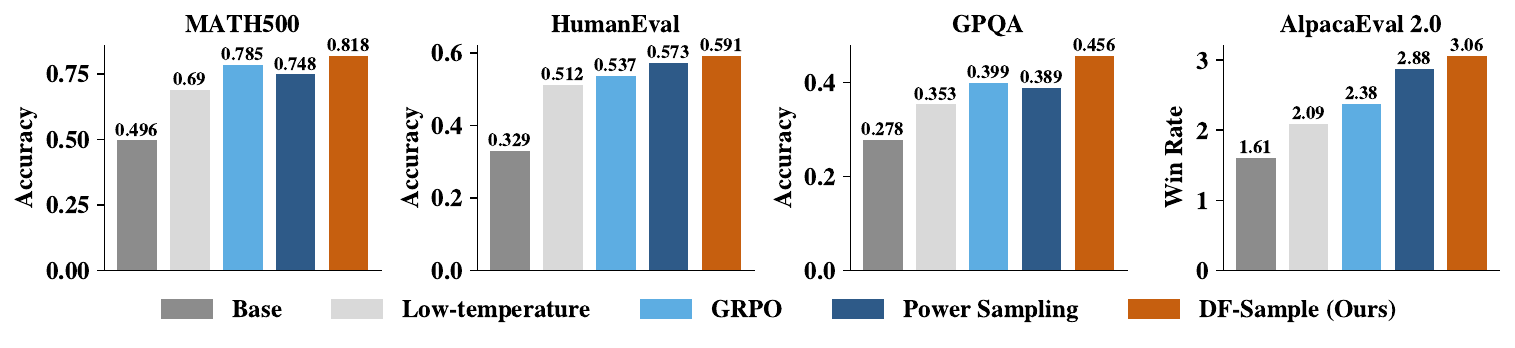}
\vspace{-0.1in}
\captionsetup{justification=raggedright,singlelinecheck=false}
\caption{DF-Sample vs.  baselines on Qwen2.5-MATH-7B using MATH500, HumanEval, GPQA,  AlpacaEval~2.0 datasets. DF-Sample achieves the best performance on different datasets. Full results are as shown in Table~\ref{tab:power_sampling}.
}
\label{fig:performace}
\vspace{-0.2in}
\end{figure*}




This raises two natural questions. 
\textbf{First}, if 
high-quality reasoning paths are already latent in the base model \cite{karan2025reasoning}, can we unlock them 
without costly RL fine-tuning
\cite{dragoi2025beyond,vafaii2025brain}? 
We argue yes. The key insight is that the problem is not one of missing knowledge but of misallocated probability, and misallocation is a sampling problem, not a training problem.
\textbf{Second}, given that a correct reasoning path already exists, how do we effectively find it? Existing approaches inspired by probabilistic sampling techniques such as MCMC~\cite{karan2025reasoning,wu2025efficiencyadaptivitydeeperlook} correct reasoning locally, one step at a time.
But local optimality at each step does not guarantee global optimality for the full trajectory. A locally plausible transition can still steer reasoning toward a globally wrong conclusion.

To address both questions, we propose \textbf{DF-Sample}: a training-free  framework that selects high-quality reasoning chains during inference by performing explicit global trajectory evaluation. Rather than making greedy or locally-guided decisions at each step, DF-Sample constructs a hierarchical reasoning tree, evaluates terminal nodes for quality, propagates those quality signals backward through the tree, and samples a final path according to a posterior that integrates generation priors with the propagated utilities.
Our contributions are as follows:

\begin{itemize}[itemsep=1pt]

\item \textbf{Decision-Flow Sampling (DF-Sample)}. A training-free inference-time framework that extracts high-quality but low-probability reasoning paths from base language models, 
improving reasoning accuracy without any parameter updates. 



\item \textbf{Global trajectory evaluation via terminal-node utility and backward propagation.} We introduce a terminal energy function that scores complete reasoning chains and a backward propagation mechanism that carries those global signals to each intermediate decision, mitigating the myopic bias of standard sampling.

\item \textbf{Extensive empirical validation.} Across three model families (Qwen2.5-Math-7B, Qwen2.5-7B, Phi-3.5-mini-instruct) and four datasets (MATH500, HumanEval, GPQA-Diamond, AlpacaEval 2.0), DF-Sample consistently outperforms base models, competitive sampling baselines, and RL-trained counterparts. On MATH500 with Qwen2.5-Math-7B, DF-Sample achieves 81.8\% accuracy, surpassing GRPO (78.5\%) and power sampling (74.8\%) as shown in Figure \ref{fig:performace}.

\end{itemize}

\section{Motivation}

Before introducing the method, we present two empirical observations about base model reasoning behavior which motivate our work.

\begin{figure}[!t]
\centering
\includegraphics[width=0.99\linewidth]{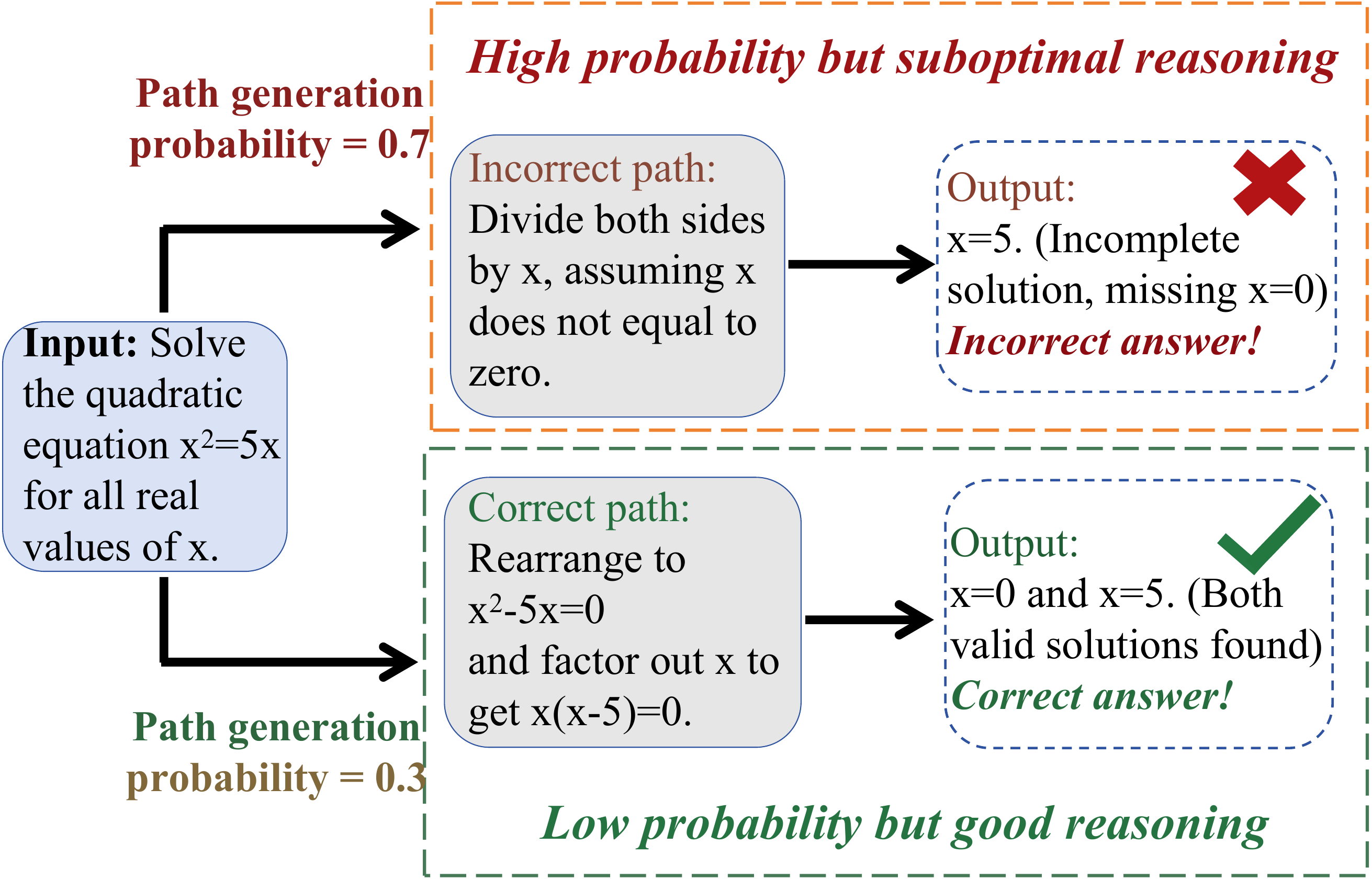}

\captionsetup{justification=raggedright,singlelinecheck=false}
\caption{Upper path (incorrect, high confidence) vs. lower path (correct, low  confidence) for quadratic equation solving problem.}
\label{fig:frequency}
\vspace{-0.2in}
\end{figure}

\textbf{Correct reasoning paths are latent, not absent.}
Building on recent evidence~\cite{karan2025reasoning, hao2025rethinking}, we observe that base models often already possess correct reasoning paths and they simply assign those paths lower probability than incorrect alternatives.
Consider the equation \emph{$x^2 = 5x$}
(Figure \ref{fig:frequency}).  
A correct solution factors both sides:
\emph{$x^2 - 5x = 0 \Rightarrow x(x-5)=0$.}, yielding both solutions.
However, the model may instead divide both sides by $x$,
implicitly assuming $x \neq 0$ and missing the solution $x = 0$.
The division path receives a higher probability because it matches common patterns and requires fewer steps, not because it is more correct. 

This is not an isolated failure mode. It reflects a systematic mismatch between generation probability and reasoning quality in base models. The implication for our method is direct: we do not need to train the model to produce better reasoning; we need a sampling procedure that looks past local probability and identifies globally superior paths.


\textbf{Local sampling optimality does not guarantee global reasoning quality.}
%
%
%
%
Given that correct reasoning paths are latent in the model's distribution, 
the question becomes how to find them. A natural candidate is MCMC-style sampling~\cite{karan2025reasoning} for correcting reasoning through local proposal-and-accept steps. However, multi-step CoT reasoning violates a key assumption that makes local corrections sufficient: the quality of a single reasoning step is not a reliable signal of the quality of the complete trajectory.

\begin{figure}[!t]
\centering
\includegraphics[width=0.99\linewidth]{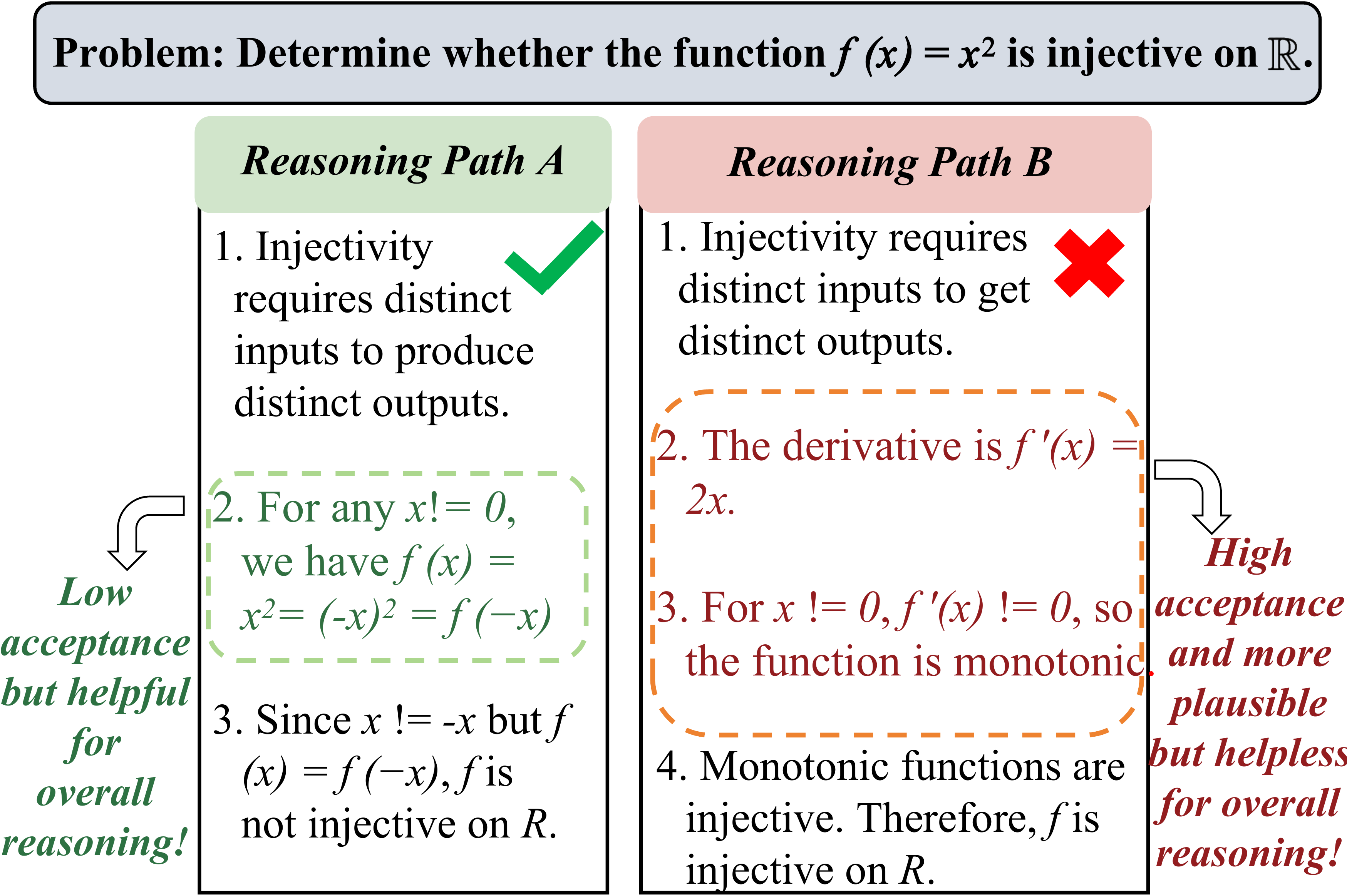}

\captionsetup{justification=raggedright,singlelinecheck=false}
\caption{Two reasoning paths for the same problem. Path B is locally plausible at each step but reaches an incorrect conclusion; Path A reasons correctly.}
\label{fig:reasoning_example}
\vspace{-0.2in}
\end{figure}

Figure~\ref{fig:reasoning_example} illustrates this with the task of 
determining 
\emph{whether $f(x)=x^2$ is injective on $\mathbb{R}$}.
Path A exploits symmetry, noting that $f(x) = f(-x)$ for $x \neq 0$, and correctly concludes $f$ is not injective. Path B introduces a derivative argument ($f'(x) = 2x$, $f'(x) \neq 0$ for $x \neq 0$, therefore monotonic, therefore injective), which is locally plausible at each step but reaches an incorrect conclusion. A sampling procedure that accepts Path B's steps because they are individually reasonable will systematically fail on problems where globally-correct reasoning is locally counterintuitive.

This motivates an approach that evaluates reasoning paths holistically, constructing a space of complete paths and selecting among them based on path-level quality rather than step-level probability.

\vspace{-0.05in}

\section{Preliminary}

\vspace{-0.05in}

Let $\mathcal{Y}$ denote a finite discrete token vocabulary, and consider sequences consisting of tokens in $\mathcal{Y}$.
A sequence of length $T \in \mathbb{Z}_{\ge 0}$ is denoted by
$\mathbf{y}_{0:T} = (y_0, y_1, \dots, y_T)$,
where each $y_t \in \mathcal{Y}$.
We use $\mathbf{y}$ to represent the full sequence.
For any time step $t$, we define the prefix (historical context) as $\mathbf{y}_{<t} = (y_0, \dots, y_{t-1})$,
and the suffix (generation of future tokens) as
$\mathbf{y}_{>t} = (y_{t+1}, \dots, y_T)$.

Under this formulation, a large language model (LLM) induces a probability distribution 
$p_\theta$ over the sequence space.
Autoregressive language models parameterize this distribution through conditional factors
$p_\theta(y_t \mid \mathbf{y}_{<t})$.
By the chain rule of probability, the joint likelihood of a sequence can be factorized as
\begin{equation}
p_\theta(\mathbf{y}_{0:T})
=
\prod_{t=0}^{T} p_\theta(y_t \mid \mathbf{y}_{<t}).
\end{equation}

Consequently, sampling a complete sequence from $p_\theta$ corresponds to 
sequentially sampling tokens according to the conditional distributions above.

\section{Proposed Method}

\begin{figure*}[!t]
\centering
\includegraphics[width=0.9\linewidth]{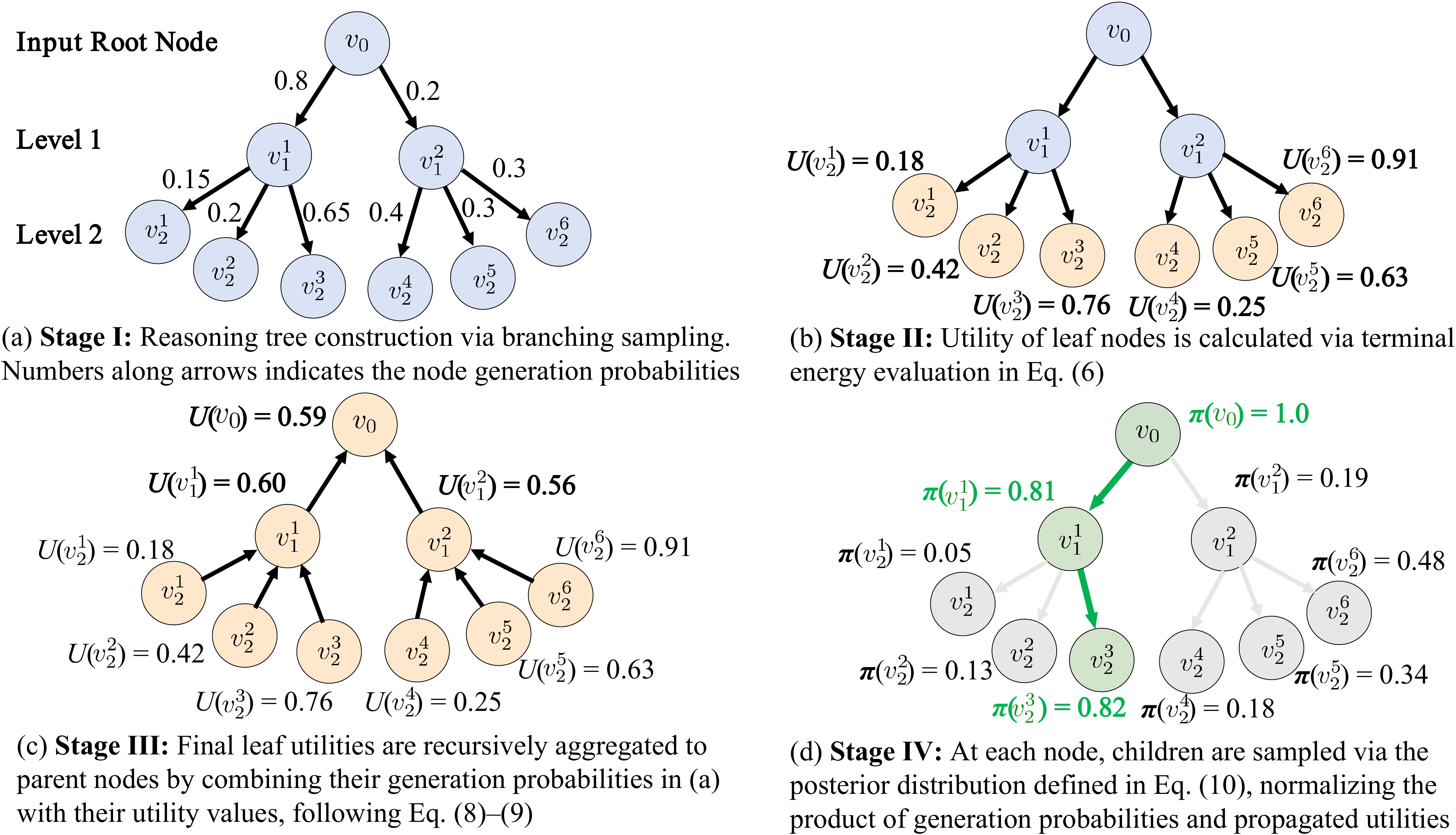}
\vspace{-0.1in}
\captionsetup{justification=raggedright,singlelinecheck=false}
\caption{An illustrative toy example of Decision-Flow sampling with branching factor $K$, showing how global reasoning strategies are selected by combining utility $U$ with prior probabilities.}
\label{fig:4_stage}
\vspace{-0.1in}
\end{figure*}

DF-Sample operates in four sequential stages, each building on the output of the previous. Specifically, \textbf{Stage I: \textit{Hierarchical Reasoning Tree Construction.}} A hierarchical reasoning tree is constructed which systematically expands candidate reasoning steps. \textbf{Stage II: \textit{Terminal Node Energy Evaluation.}} We assign each final node an utility score based on the quality of its corresponding final reasoning output.
\textbf{Stage III: \textit{ Decision-Flow Backward Propagation.}} Terminal evaluation utility are propagated backward through the tree, enabling intermediate reasoning states to incorporate and obtain globally node-level utilities or utility. \textbf{Stage IV: \textit{ Posterior Path Selection.}} A reasoning path is selected according to a posterior decision rule that integrates prior generation tendencies with the propagated utility, and the final answer is derived from the selected chain. A toy example is shown in Figure~\ref{fig:4_stage} to illustrate the four stages.


\subsection{Hierarchical Reasoning Tree Construction}

Given an input problem $q$, we construct a reasoning tree $\mathcal{T}$ of depth $L$ by hierarchical sampling. At each node
of depth $\ell-1$, we independently sample $K$ candidate extensions from the language model, forming total child nodes $\mathcal{C}(v_{\ell-1}) = \{v^1_\ell, v^2_\ell, \ldots, v^{dK}_\ell\}$ ($d$ denotes the number of nodes at depth $\ell-1$). The $i$-th child $v^i_\ell$ at depth $\ell$ corresponds to one reasoning step $s_{v^i_\ell}$ generated conditioned on the problem and the path from the root to the parent of $v_{\ell}^i$.

For each node $v$, we record both the generated reasoning text $s_v$ and the accumulated log-probability along the path from the root:
\begin{equation}
    \log p(s_v \mid q,\, s_{<v}) = \sum_{t=1}^{|s_v|} \log p(w_t \mid q,\, s_{<v},\, w_{<t})
\end{equation}
The full tree is constructed before any selection is made, enabling subsequent global evaluation over the entire reasoning space, a key departure from step-wise decoding strategies.

\begin{figure}[!t]
\centering
\includegraphics[width=0.99\linewidth]{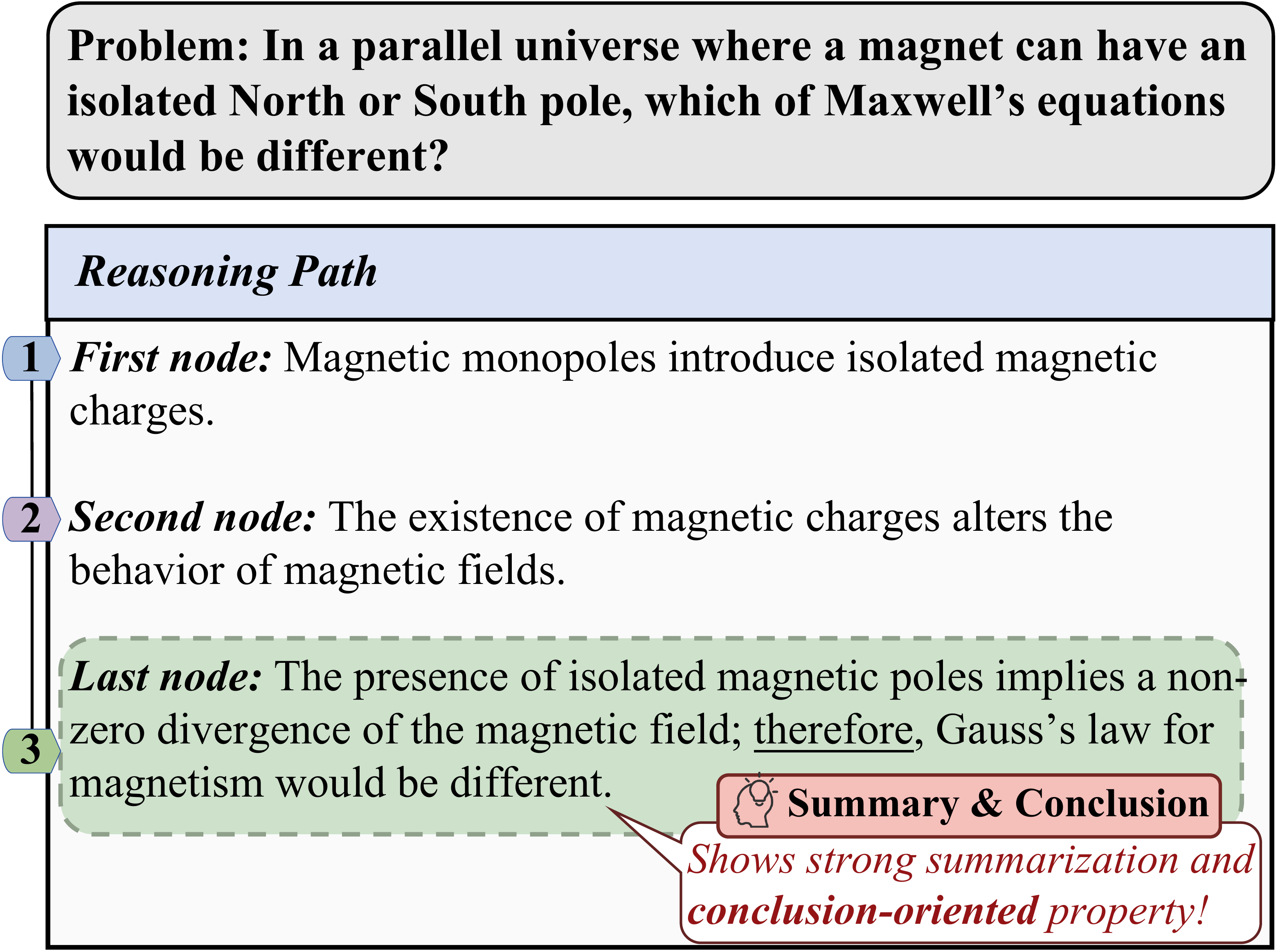}

\captionsetup{justification=raggedright,singlelinecheck=false}
\caption{Final node exhibits strong summarization and conclusion-oriented characteristics.}
\label{fig:final_node_box_example}
\vspace{-0.2in}
\end{figure}

\subsection{Terminal Node Energy Evaluation}

We evaluate the quality of complete reasoning chains by scoring their terminal nodes. This design is motivated by a consistent empirical observation as shown in Figure~\ref{fig:final_node_box_example}: the final reasoning step typically exhibits strong summarization and conclusion-oriented properties, condensing the chain's outcome into a concise statement. Evaluating the terminal node therefore serves as an efficient proxy for trajectory quality.

Define a terminal energy function over leaf nodes:
\begin{equation}
E(\tau)
\;=\;
- \alpha \cdot
\frac{\log p\!\left(s_{v_{L}} \mid q, s_{<v_{L}}\right)}{|s_{v_{L}}|} + R(s_{v_{L}}),
\label{eq6}
\end{equation}
where $\alpha$ is a temperature coefficient controlling the influence of model likelihood, 
$|s_{v_{L}}|$ denotes the number of tokens
of the final reasoning step, and $R(s_{v_{L}})$ is a quality score for the final output.
A lower 
$R(s_{v_{L}})$ indicates higher output quality and a lower energy $E$ indicates a higher-quality path.

To evaluate the significance of leaf nodes, we define
the terminal utility of each leaf node $v_{L}$ as $U(v_{L})$. This metric facilitates the propagation of path-level evaluations back through the entire reasoning tree and is formulated as
\begin{equation}
U(v_{L}) = \exp\!\left(-E(v_{L})\right)
\end{equation}
where higher $U(v_{L})$ indicates greater terminal utility. These leaf-level scores are then propagated backward through the three in Stage III.

\subsection{Decision-Flow Backward Propagation}

Having scored all terminal nodes, we propagrate utility signals backward through the tree in a bottom-up pass.
For any non-leaf node $v_{\ell-1}^t$, its utility is defined  as the prior-weighted expected utility of its $K$ children, a recursive aggregation that carries trajectory-level quality information to each intermediate decision state: 
\begin{equation}
U(v_{\ell-1}^t) = \sum_{i=tK+1}^{(t+1)K} p_{\text{prior}}(v_{\ell}^i \mid v_{\ell-1}^t)\, U(v_{\ell}^i),
\end{equation}
where $v_{\ell}^i$ represents the $i$-th node at Level $\ell$ (the ($i$ \% $K$)-th child of $v_{\ell-1}^t$).  

The prior transition probability $p_{\text{prior}}(v_{\ell}^i \mid v_{\ell-1}^t)$ is defined by normalizing the language model’s generation probabilities over the $K$ sampled children at each node:
\begin{equation}
\resizebox{\linewidth}{!}{$
p_{\text{prior}}(v_{\ell}^i \mid v_{\ell-1}^t)
=
\frac{\exp\!\left(\log p(s_{v_{\ell}^i} \mid v_{\ell-1}^t, s_{<v_{\ell-1}^t})\right)}
{\sum_{j=tK+1}^{(t+1)K} \exp\!\left(\log p(s_{v_{\ell}^j} \mid v_{\ell-1}^t, s_{<v_{\ell-1}^t})\right)}
$}
\end{equation}
This backward propagation mechanism is what distinguishes DF-Sample from locally-guided methods: every intermediate node accumulates information about the global quality of the paths it can reach, rather than making decisions based solely on the local next-step probability distribution.

\subsection{Posterior Path Selection}

With utilities computed for all nodes, we define a posterior selection policy that integrates generation priors with propagated utilities. 
For node $v_{\ell-1}^t$, the probability of selecting its $i$-th child $v_{\ell}^i$ is defined as:
\begin{equation}
\pi^{*}(v_{\ell}^i \mid v_{\ell-1}^t)
=
\frac{p_{\text{prior}}(v_{\ell}^i \mid v_{\ell-1}^t)\, U(v_{\ell}^i)}
{\sum_{j=tK+1}^{(t+1)K} p_{\text{prior}}(v_{\ell}^j \mid v_{\ell-1}^t)\, U(v_{\ell}^j)}.
\end{equation}

Starting from the root node, we recursively sample children according to the
posterior distribution
$\pi^{*}(\cdot \mid \cdot)$ until reaching a final leaf node, yielding
a complete reasoning path $\hat{\tau}$. If $\hat{\tau}$ contains an explicit final answer, we output it directly; otherwise, $\hat{\tau}$ is appended to the original prompt and used as context for generating a final answer.

The posterior selection rule reduces the myopic bias of purely local accept-reject strategies by weighting each child not only by how likely it is but by how good the paths it leads to tend to be. A locally improbable step that consistently leads to correct conclusions will be favored over a locally probable step that leads to wrong ones.

\begin{algorithm}[t!]
\small
\caption{\small Decision-Flow Sample (DF-Sample)}
\label{alg:df_sample}
\begin{algorithmic}[1]

\State \textbf{Input:} Question $q$, model $\mathcal{M}$, branching factor $K$, total depth $N$, block size $B$, temperature $\alpha$
\State \textbf{Output:} Selected reasoning chain $\hat{\tau}$

\State Initialize context $c \gets q$ (question with prompt)
\State Initialize final chain $\hat{\tau} \gets \emptyset$
\State $t \gets 0$

\While{$t < N$}

\State $L \gets \min(B, N-t)$
\State Initialize reasoning tree $\mathcal{T}$ with root node $v_0$ (using context $c$)

\State {\color{blue}\textit{// Phase 1: Hierarchical Reasoning Tree Construction}}
\For{$\ell = 1$ \textbf{to} $L$}
  \For{each node $v$ at depth $\ell-1$}
    
    \State Create leaf nodes $ \{v_{\ell}^1, \dots, v_{\ell}^K\}$
    \State Sample $K$ candidate reasoning steps $\{s_{v_{\ell}^1},\ldots,s_{v_{\ell}^K}\}$ from $\mathcal{M}$
    \State Add leaves to $\mathcal{T}$
  \EndFor
\EndFor

\State {\color{blue}\textit{// Phase 2: Terminal Node Energy Evaluation}}
\For{each leaf node $v_{L}$}
  \State Compute terminal energy $E(v_{L})$ using Eq.~(3)
  \State $U(v_{L}) \gets \exp(-E(v_{L}))$
\EndFor

\State {\color{blue}\textit{// Phase 3: Decision-Flow Backward Propagation}}
\For{$\ell = L-1$ \textbf{downto} $0$}
  \For{each node $v$ at depth $\ell$}
    \State $U(v_{\ell}) = \sum_i p_{\text{prior}}(v_{\ell+1}^i \mid v_{\ell})\, U(v_{\ell+1}^i)$
  \EndFor
\EndFor

\State {\color{blue}\textit{// Phase 4: Posterior Path Selection}}
\State $\tau_b \gets \emptyset$, $v \gets v_0$
\While{$\ell<L$ }
  \State Sample $v_{\ell}^i$ with probability
  \[
  \pi^*(v_{\ell}^i \mid v)
  \propto
  p_{\text{prior}}(v_{\ell}^i \mid v)\,U(v_{\ell}^i)
  \]
  \State Append $s_{v_{\ell}^i}$ to $\tau_b$
  \State $v \gets v_{\ell}^i$
\EndWhile

\State Append $\tau_b$ to $\hat{\tau}$
\State Update context $c \gets (q,\hat{\tau})$
\State $t \gets t + L$

\EndWhile

\State \textbf{return} $\hat{\tau}$

\end{algorithmic}
\end{algorithm}

\begin{table*}[!t]
\centering
\caption{
Performance comparison of DF-Sample and baselines on MATH500, HumanEval, GPQA-Diamond, and AlpacaEval 2.0. \textbf{Bold} denotes the best performance. DF-Sample outperforms all baselines in most settings.
}
\label{tab:power_sampling}
\resizebox{0.85\linewidth}{!}{
\begin{tabular}{lcccc}
\hline
 & \textbf{MATH500} & \textbf{HumanEval} & \textbf{GPQA-Diamond} & \textbf{AlpacaEval~2.0} \\
\hline

\textbf{Qwen2.5-Math-7B} \\
Base                 & 0.496 & 0.329 & 0.278 & 1.61 \\
Low-temperature      & 0.690 & 0.512 & 0.353 & 2.09 \\
GRPO (MATH)          & 0.785 & 0.537 & 0.399 & 2.38 \\
Power Sampling       & 0.748 & 0.573 & 0.389 & 2.88 \\
\hline
\textbf{DF-Sample (Ours)}& \textbf{0.818} & \textbf{0.591} & \textbf{0.456} & \textbf{3.06} \\
\hline

\textbf{Qwen2.5-7B} \\
Base                 & 0.498 & 0.329 & 0.278 & 7.05 \\
Low-temperature      & 0.628 & 0.524 & 0.303 & 5.29 \\
GRPO (MATH)          & \textbf{0.740} & 0.561 & 0.354 & 7.62 \\
Power Sampling       & 0.706 & 0.622 & 0.318 & 8.59 \\
\hline
\textbf{DF-Sample (Ours)}& 0.736 & \textbf{0.646} & \textbf{0.359} & \textbf{9.19} \\
\hline

\textbf{Phi-3.5-mini-instruct} \\
Base                 & 0.400 & 0.213 & 0.273 & 14.82 \\
Low-temperature      & 0.478 & 0.585 & 0.293 & \textbf{18.15} \\
GRPO (MATH)          & 0.406 & 0.134 & 0.359 & 16.74 \\
Power Sampling & 0.508 & \textbf{0.732} & 0.364 & 17.65 \\

\hline
\textbf{DF-Sample (Ours)}& \textbf{0.544} & 0.665 & \textbf{0.394} & 17.89\\
\hline

\end{tabular}
}
\vspace{-0.15in}
\end{table*}

\subsection{Efficient Inference via Block-wise Sampling}

The algorithmic pipeline involves two key design considerations. First, before constructing the reasoning tree, the model estimates the required number of reasoning steps $N$, which determines the tree depth needed to produce a complete reasoning trajectory. Second, as $N$ grows, the number of candidate nodes expands exponentially, incurring substantial computational overhead.

To mitigate this complexity, we introduce a block-wise sampling strategy.
Specifically, when $N$ exceeds a predefined threshold $B$, we first construct and evaluate the reasoning tree for the initial block of steps and apply Decision-Flow selection to identify the best partial trajectory.
The selected partial reasoning path is then appended to the original problem 
as context for generating subsequent reasoning blocks.
This process repeats until the full reasoning depth $N$ is reached. 
The Pseudocode of Decision-Flow Sampling is shown in \textbf{Algorithm 1}.





\section{Experiments}


\subsection{Setup}

\textbf{Datasets.} We evaluate on four benchmarks spanning maths, codes, science, and general helpfulness:
\textbf{MATH500:} 500 competition-level mathematics problems from the MATH dataset
\cite{lightman2023letsverifystepstep}, covering
algebra, geometry, and number theory. 
\textbf{HumanEval:}
164 hand-written programming tasks~\cite{chen2021evaluatinglargelanguagemodels}. 
Solutions are evaluated by executing unit tests and a problem is correct only if all tests pass.
\textbf{GPQA-Diamond:} 198 graduate-level multiple-choice questions in physics, chemistry, and biology~\cite{rein2024gpqa}, the most difficult split of GPQA.
\textbf{AlpacaEval 2.0:} 805 open-ended instruction-following prompts~\cite{dubois2024length}.
Responses are judged by GPT-4-Turbo and reported as a length-normalized win rate.

\textbf{Models and Baselines.}
We evaluate on three base model families: Qwen2.5-Math-7B, Qwen2.5-7B, and Phi-3.5-mini-instruct. We compare against four methods:  \textbf{Base model:} Standard greedy decoding from the base model, with no sampling modification; \textbf{Low-temperature sampling:} Exponentiated conditional next-token distributions at each step\cite{wang2020contextualtemperaturelanguagemodeling}, which sharpens the output distribution without constructing a tree;
\textbf{GRPO:} Group Relative Policy Optimization~\cite{shao2025spurious}, an RL-based fine-tuning method; \textbf{Power sampling: } improves reasoning by sampling many candidate reasoning segments from base model and selecting better replacement segment via MCMC~\cite{karan2025reasoning}.
All baseline results are taken directly from the corresponding papers \cite{karan2025reasoning}.

\textbf{Implementations.}
In our implementation, following prior work~\cite{karan2025reasoning}, we set $\alpha=4.0$ to encourage the model to produce outputs with higher confidence. We use $K=3$ and $B=3$ in all experiments. All other hyperparameters are kept consistent with the Power Sampling baseline. The value of 
$R$ is evaluated using GPT-4o. All experiments are run on 2 NVIDIA A6000 GPUs.

\subsection{Main Results}


DF-Sample consistently improves over the base model across all settings and outperforms the training-free baseline power sampling in most configurations. The gains are particularly striking on GPQA-Diamond, where DF-Sample with Qwen2.5-Math-7B achieves 45.6\% compared to GRPO's 39.9\%. This is notable because GPQA-Diamond problems require multi-step scientific reasoning where globally-correct paths are most likely to be locally counterintuitive — precisely the regime where global trajectory evaluation provides the largest advantage over step-wise methods.
On MATH500, DF-Sample achieves 81.8\% with Qwen2.5-Math-7B, exceeding GRPO (78.5\%) by 3.3 percentage points without any parameter updates. On AlpacaEval 2.0, DF-Sample's win rate improvements generalize beyond verifiable reasoning tasks, suggesting that the latent path hypothesis extends to general instruction following.

\begin{figure*}[!t]
\centering

\begin{minipage}{0.48\linewidth}
\centering
\includegraphics[width=\linewidth]{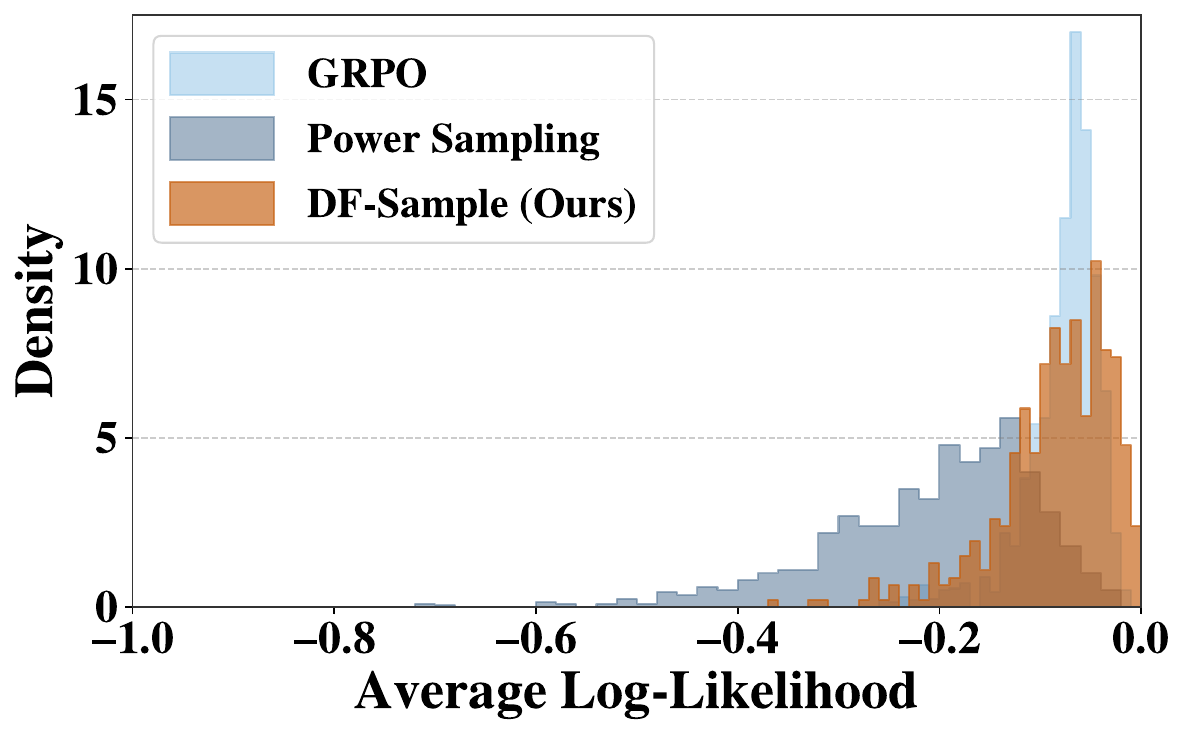}
\vspace{-0.3in}
\caption{The average log-likelihood of Ours, power sampling, and GRPO responses over MATH500.}
\label{fig:avg_like}
\end{minipage}
\hfill
\begin{minipage}{0.48\linewidth}
\centering
\includegraphics[width=\linewidth]{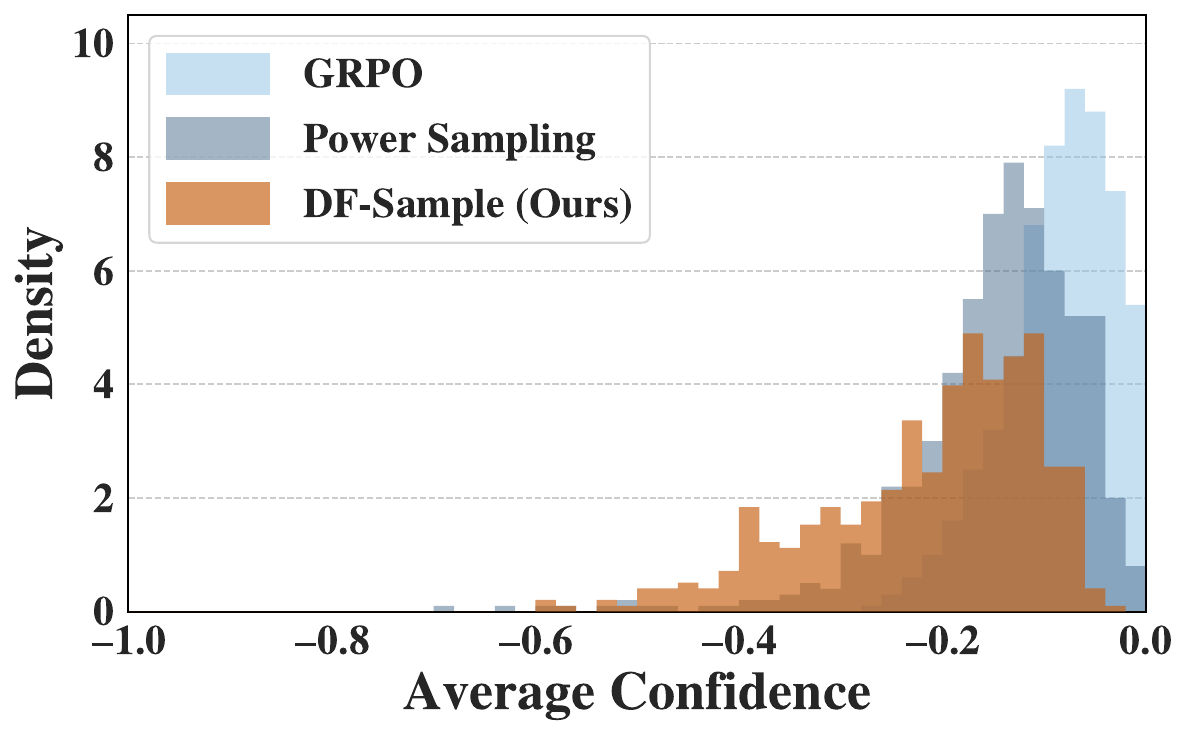}
\vspace{-0.3in}
\caption{The average token confidence of Ours, power sampling, and GRPO responses over MATH500.}
\label{fig:avg_conf}
\end{minipage}

\end{figure*}

\subsection{Analysis}

\textbf{Token likelihood distribution.} 
Figure \ref{fig:avg_like} shows the histogram of length-normalized sequence log-likelihoods for DF-Sample, GRPO, and power sampling on MATH500. GRPO concentrates mass in a narrow high-likelihood region, consistent with distribution sharpening. DF-Sample is biased toward higher likelihood but retains a broader spread than GRPO, indicating it accesses a wider region of the base model's distribution while still favoring higher-quality paths.

\textbf{Token-level confidence.}
Figure \ref{fig:avg_conf} compares the distribution of token-level confidence, defined as the average negative confidence of next-token predictions.
\begin{equation}
\resizebox{0.85\linewidth}{!}{$
\mathrm{Conf}(x_{0:T}) = \frac{1}{T+1} 
\sum_{t=0}^{T} 
\sum_{x \in \mathcal{X}} 
p(x \mid x_{<t}) \log p(x \mid x_{<t}).
$}
\end{equation}
GRPO responses concentrate near the highest-confidence region, indicating that the generated tokens are typically produced in locally confident contexts under the base model. Power sampling shows a somewhat broader distribution of confidence values. In contrast, DF-Sample shows the largest spread and extends furthest into low-confidence regions. This reveals that DF-Sample is selecting paths through parts of the reasoning space where the model is locally uncertain but globally correct, precisely the latent paths the method is designed to recover.


\begin{figure}[!t]
\centering
\includegraphics[width=0.9\linewidth]{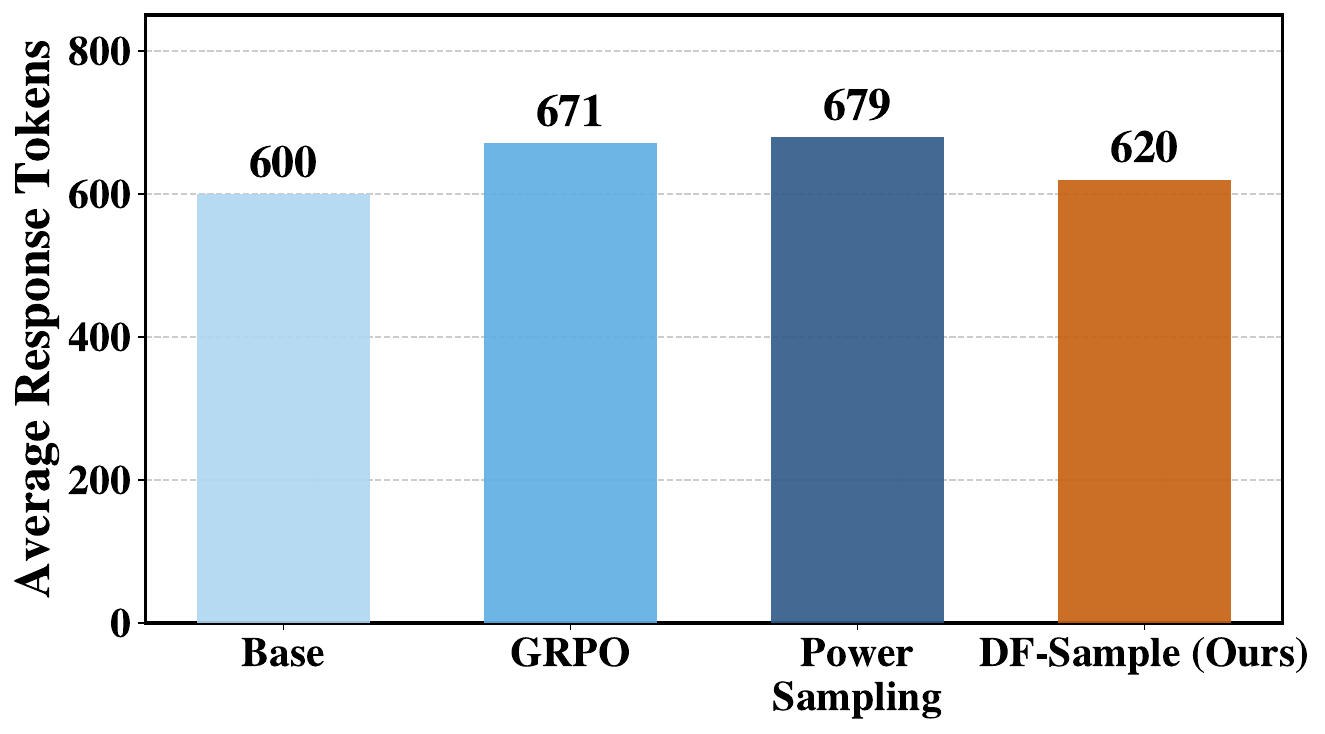}
\vspace{-0.1in}
\captionsetup{justification=raggedright,singlelinecheck=false}
\caption{DF-Sample (Ours) vs. baselines on average generated tokens with Qwen2.5-Math-7B on MATH500 dataset. 
}
\label{fig:generate_token}
\vspace{-0.2in}
\end{figure}

\textbf{Average response lengths and latency.}
Figure~\ref{fig:generate_token} compares the average number of tokens generated for DF-Sample, power sampling, and GRPO relative to the base model on the MATH500 dataset. Power sampling produces the longest responses, followed by GRPO, indicating that these methods tend to generate longer reasoning trajectories. In contrast, DF-Sample generates responses with lengths of about 619 comparable to the base model of about 600. This suggests that DF-Sample tends to select relatively shorter reasoning trajectories while still preserving relatively correct answers. On the MATH500 dataset with Qwen2.5-Math-7B, power sampling requires about 340 seconds per question, while DF-Sample takes around 384 seconds. Despite the slightly longer inference time, DF-Sample achieves 81.8\% accuracy, outperforming power sampling by 7.0\%.

\begin{figure}[!t]
\centering
\includegraphics[width=0.9\linewidth]{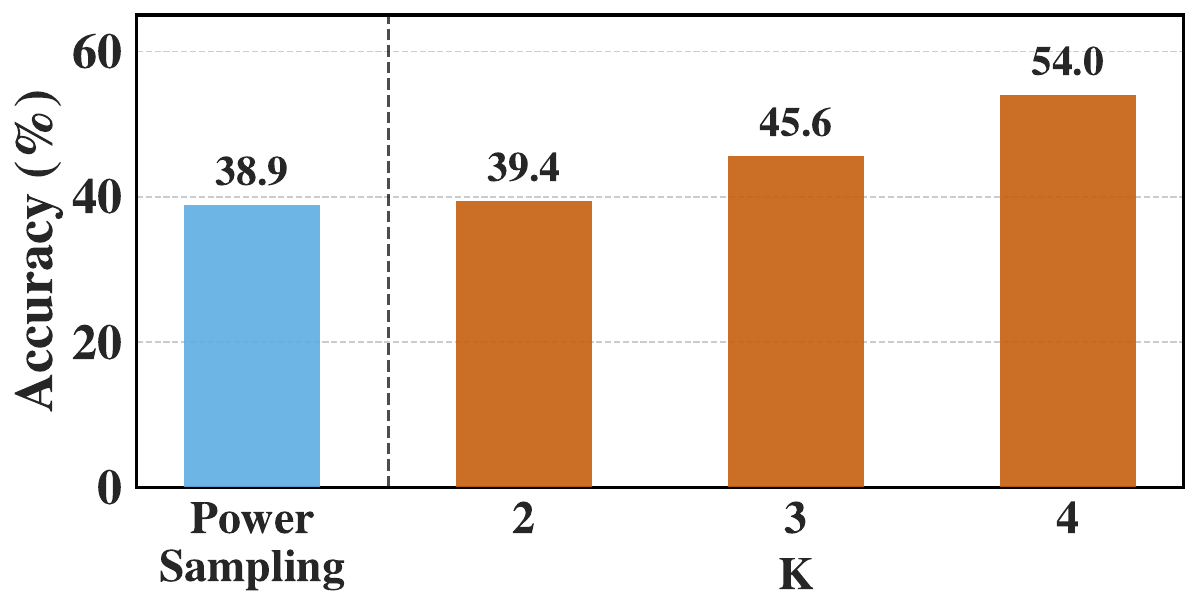}
\vspace{-0.1in}
\captionsetup{justification=raggedright,singlelinecheck=false}
\caption{Accuracy comparison of DF-Sample (with different $K$) and Power-Sampling with Qwen2.5-Math-7B on GPQA-Diamond.
}
\label{fig:effect K}
\vspace{-0.2in}
\end{figure}

\subsection{Ablation Study}

\textbf{Impact of $K$.}
Figure~\ref{fig:effect K} shows the performance of DF-Sample on GPQA with Qwen2.5-Math-7B under different $K$. 
The model is more likely to find correct reasoning paths with larger $K$, achieving higher accuracy. 
For example, the accuracy is only about 1\% higher than power sampling when $K=2$, while when $K=4$ it improves by about 18\%. 
However, larger $K$ increases inference latency, so we set $K=3$ to balance accuracy and efficiency.

\begin{figure}[!t]
\centering
\includegraphics[width=0.95\linewidth]{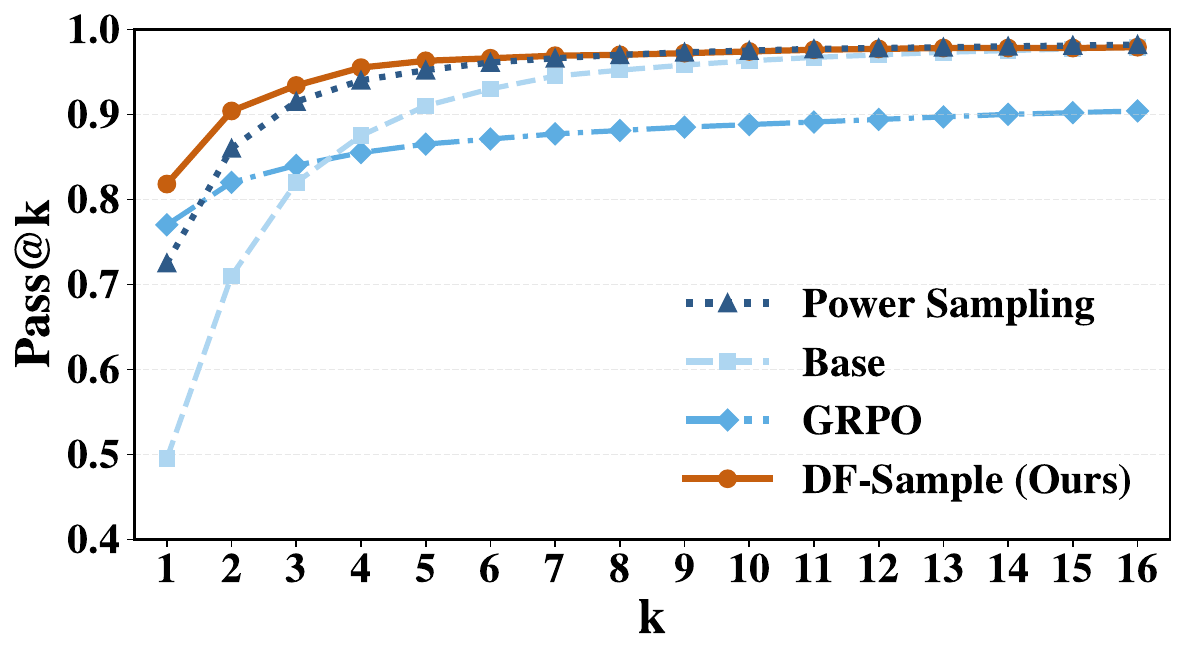}
\vspace{-0.1in}
\captionsetup{justification=raggedright,singlelinecheck=false}
\caption{DF-Sample (Ours) vs. baselines on relative Pass@k accuracy to base model (Qwen2.5-Math-
7B) on MATH500 dataset.}
\label{fig:7}
\vspace{-0.15in}
\end{figure}

\begin{table}[!t]
\centering
\caption{Ablation study on $\alpha$ on the GPQA-Diamond dataset using Qwen2.5-Math-7B.}
\label{tab:alpha}
\begin{tabular}{c  c}
\toprule
\textbf{Value of $\alpha$} & \textbf{Accuracy (\%)} \\
\midrule
100 & 30.3 \\
10 & 40.9 \\
5 & 48.0 \\
4 & 45.9 \\
1 & 27.8 \\
\bottomrule
\vspace{-0.3in}
\end{tabular}
\end{table}

\textbf{Pass@k.} 
Figure \ref{fig:7} compares the pass@k accuracy of DF-Sample and baselines on the MATH500 dataset. 
DF-Sample achieves the highest pass@1 and pass@2 accuracy, with advantages particularly pronounced in the low-k regime.
As $k$ increases, the performance of the two methods becomes comparable, with power sampling slightly surpassing DF-Sample at a few larger $k$ values. Nevertheless, the differences remain small, indicating that DF-Sample is highly effective at extracting correct reasoning paths, especially when only a small number of samples are available.



\textbf{Impact of temperature coefficient $\alpha$.}
Table~\ref{tab:alpha} reports the ablation on the temperature coefficient $\alpha$ on GPQA with Qwen2.5-Math-7B. 
When $\alpha$ is too large, sampling overly favors high-probability reasoning paths and degrades accuracy. 
When $\alpha$ is too small, the distribution becomes overly flat and lacks sufficient discrimination among candidate reasoning paths. 
We observe the best performance when $\alpha>1$ with moderate values (around $4$ or $5$).

\vspace{1mm}

\section{Related works}


\textbf{RL for LLM reasoning.}
RL-based approaches improve reasoning through post-training. 
Early work applies RLHF \cite{dong2024rlhf} to align outputs with human preferences, while more recent methods adopt reinforcement learning with verifiable rewards (RLVR) that directly optimize task correctness using automated verifiers \cite{lambert2024tulu,guo2025deepseek,zeng2025simplerl}. 
GRPO \cite{shao2025spurious} is a representative RLVR method that improves reasoning via group-relative policy updates. 
Despite strong results, such approaches require expensive fine-tuning and may reduce output diversity \cite{shao2025spurious,he2025rewarding}. 
In contrast, DF-Sample achieves similar improvements without parameter updates.

\textbf{Inference-Time Reasoning Search.}
Another line of work improves reasoning by allocating additional inference-time computation.
Methods such as Tree of Thoughts \cite{yao2023treethoughtsdeliberateproblem}, HyperTree \cite{gui2025hypertreeplanningenhancingllm}, and ReST-MCTS \cite{zhang2024restmctsllmselftrainingprocess} explore tree-structured reasoning search, while adaptive branching dynamically allocates compute during reasoning \cite{inoue2025widerdeeperscalingllm}.
Power Sampling \cite{karan2025reasoning} reshapes reasoning distributions via MCMC-style sampling.
In contrast, DF-Sample evaluates reasoning trajectories using global path-level utilities.


\textbf{Sampling-Based Methods.}
Sampling-based approaches such as MCMC and Generative Flow Networks (GFlowNets) \cite{bengio2021flow} reshape or approximate a model’s output distribution to favor high-reward trajectories \cite{neal2001annealed}. 
For example, QUEST \cite{faria2024quest} applies a Metropolis–Hastings variant to iteratively resample model outputs, while annealed sampling \cite{karan2025reasoning} sharpens the output distribution via temperature scheduling. 
The Decision Flow framework \cite{chertkov2025samplingdecisions} formulates trajectory sampling using a GFlowNet-style approach with backward regression over sequence graphs. 
Building on this idea, DF-Sample applies decision-flow-style evaluation to reasoning trajectories, improving reasoning accuracy in LLMs.

\vspace{1.5mm}

\section{Conclusion}
\vspace{1.5mm}

In this paper, we propose \textbf{Decision-Flow Sampling (DF-Sample)}, a training-free inference-time sampling framework for improving reasoning performance in LLMs. 
Unlike conventional decoding strategies that rely on local step-wise probabilities, DF-Sample constructs a hierarchical reasoning tree and performs global trajectory-level evaluation through terminal-node utility estimation and backward utility propagation. 
By integrating generation probabilities with propagated utilities, our method effectively identifies high-quality yet low-probability reasoning paths that are often overlooked during standard decoding. 
Extensive experiments across diverse reasoning benchmarks demonstrate that DF-Sample consistently improves over base models and competitive sampling baselines, and achieves performance comparable to RL-based approaches without any parameter updates. 

\section*{Limitations}

DF-Sample operates purely at inference time and does not modify model parameters. While this design enables training-free improvements, it also raises the question of whether the reasoning patterns discovered during sampling could be distilled back into the model through training or fine-tuning. For example, future work could explore integrating DF-Sample-style trajectory selection with learning-based approaches, allowing models to internalize effective reasoning strategies rather than relying solely on search during inference.




\bibliography{custom}

\appendix



\end{document}